\documentclass{article} 
\usepackage{iclr2025_conference,times}
\usepackage{natbib}

\usepackage{amsmath,amsfonts,bm}

\def\eqref#1{equation~\ref{#1}}

\def\1{\bm{1}}

\def\ra{{\textnormal{a}}}

\def\rx{{\textnormal{x}}}

\def\rva{{\mathbf{a}}}

\def\erva{{\textnormal{a}}}

\def\ervx{{\textnormal{x}}}

\def\rmA{{\mathbf{A}}}

\def\vmu{{\bm{\mu}}}
\def\vtheta{{\bm{\theta}}}
\def\va{{\bm{a}}}

\def\ve{{\bm{e}}}

\def\vx{{\bm{x}}}

\def\eva{{a}}

\def\mA{{\bm{A}}}

\def\mH{{\bm{H}}}
\def\mI{{\bm{I}}}
\def\mJ{{\bm{J}}}

\def\mX{{\bm{X}}}

\def\mSigma{{\bm{\Sigma}}}

\DeclareMathAlphabet{\mathsfit}{\encodingdefault}{\sfdefault}{m}{sl}
\SetMathAlphabet{\mathsfit}{bold}{\encodingdefault}{\sfdefault}{bx}{n}
\newcommand{\tens}[1]{\bm{\mathsfit{#1}}}
\def\tA{{\tens{A}}}

\def\tX{{\tens{X}}}

\def\gG{{\mathcal{G}}}

\def\sA{{\mathbb{A}}}
\def\sB{{\mathbb{B}}}

\def\sS{{\mathbb{S}}}

\def\emA{{A}}

\newcommand{\etens}[1]{\mathsfit{#1}}

\def\etA{{\etens{A}}}

\newcommand{\E}{\mathbb{E}}

\newcommand{\R}{\mathbb{R}}

\newcommand{\KL}{D_{\mathrm{KL}}}
\newcommand{\Var}{\mathrm{Var}}

\newcommand{\Cov}{\mathrm{Cov}}

\newcommand{\normltwo}{L^2}
\newcommand{\normlp}{L^p}

\newcommand{\parents}{Pa} 

\usepackage{hyperref}
\usepackage{url}
\usepackage{graphicx}
\usepackage{booktabs}
\usepackage{siunitx} 
\usepackage{booktabs}
\usepackage{threeparttable}
\usepackage[utf8]{inputenc}
\usepackage{amsmath, amssymb}
\usepackage{graphicx}
\usepackage{enumitem}
\usepackage{hyperref}

\title{ChameleonLLM: Batch-Aware Dynamic Low-Rank Adaptation via Inference-Time Clusters}

\author{Kamer Ali Yuksel \& Hassan Sawaf \\
        aiXplain Inc., San Jose, CA, USA \\
  \texttt{\{kamer, hassan\}@aixplain.com} \\}

\iclrfinalcopy 
\begin{document}

\maketitle

\begin{abstract}
Recent advances in large language models (LLMs) have shown remarkable performance across diverse tasks. However, these models are typically deployed with fixed weights, which limits their ability to adapt dynamically to the variability inherent in real-world data during inference. This paper introduces ChameleonLLM, a novel framework that enables inference-time adaptation of LLMs by leveraging batch-aware clustering and on-the-fly generation of low-rank updates. Unlike traditional fine-tuning approaches such as Low-Rank Adaptation (LoRA) or methods that rely on a fixed set of pre-learned uniforms (changeable masks), our method dynamically generates adaptive modifications to the decoder weights based on the aggregated statistics of clustered batches. By intelligently grouping similar inputs and computing context-aware low-rank updates via a hyper-network, ChameleonLLM achieves significant performance gains, outperforming conventional LoRA methods while eliminating the overhead of maintaining multiple expert models. Our experiments highlight the potential of our approach to serve as a versatile and highly adaptive solution for language model inference. ChameleonLLM is \href{https://anonymous.4open.science/r/ChamaleonLLM/}{open-sourced} to ensure the reproducibility of our experiments.
\end{abstract}

\section{Introduction}

Large language models have revolutionized natural language processing by demonstrating unprecedented text generation, summarization, translation, and beyond capabilities. Despite their impressive performance, a persistent limitation is their static nature during inference: once deployed, the weights of these models remain fixed regardless of the variability in input data. This static configuration can lead to suboptimal performance when encountering novel or contextually distinct inputs. Traditionally, fine-tuning methods such as Low-Rank Adaptation (LoRA) have been employed during training to inject task-specific updates into the model parameters in a computationally efficient manner \citep{hu2021lora}. LoRA achieves this by freezing most of the pre-trained model’s parameters and introducing trainable low-rank matrices that serve as corrections. However, even with these efficient updates, the low-rank modifications remain static during inference, meaning that the model cannot adjust to nuances in the input data it receives on-the-fly. 

ChameleonLLM is designed to address this limitation by enabling the model to adapt its decoder weights during inference based on the structure and statistics of the input batch. Inputs are grouped into clusters based on their semantic and syntactic similarities. By leveraging precomputed token embeddings, the inference engine identifies coherent groups within each batch, ensuring that similar examples are processed together. Rather than relying on pre-learned and fixed uniforms (changeable masks) or static LoRA updates, ChameleonLLM employs a hyper-network \citep{ha2016hypernetworks} to generate low-rank adaptation parameters in real-time. This dynamic generation is based on the aggregated statistics of the clustered batch, allowing the model to tailor its adaptations to the prevailing context. This approach not only removes the necessity for multiple expert models or a vast wardrobe of pre-stored masks but also leverages the collective context of the batch, leading to enhanced performance across a range of tasks. The primary contributions of this work are three-fold: (1) We introduce a method to dynamically generate low-rank updates based on batch statistics, resulting in a self-adaptive model during inference. (2)  We employ a clustering-based on normalized token embeddings that groups similar inputs, ensuring context-aware adaptation. (3) We provide experimental results demonstrating the superiority of ChameleonLLM over traditional LoRA fine-tuning.

\section{Related Work}

LoRA has emerged as a popular method for fine-tuning large pre-trained models with minimal computational resources. The fundamental idea behind LoRA is to freeze the pre-trained weights and introduce a pair of trainable, low-rank matrices that approximate the necessary updates for a given task. By operating in a low-dimensional subspace, LoRA significantly reduces the number of trainable parameters while enabling effective adaptation. Despite its success, standard LoRA is inherently static during inference. Once the low-rank updates are learned, they are fixed and applied uniformly, regardless of the heterogeneity of the incoming data. This static nature can limit the model's responsiveness to dynamic input distribution or context changes. Several methods have been proposed that explore dynamic and parameter-efficient adaptation from various perspectives. For instance, \citet{chen2024multiplicative} introduced a multiplicative sparse factorization technique that leverages the low-rank structure in fine-tuning, showing that a single adjustment may not be sufficient for robust performance. Similarly, \citet{kim2024single} demonstrated that a single linear layer can yield task-adapted low-rank matrices, highlighting the potential for dynamic low-rank adaptations. \citet{xiao2023task} with task-agnostic low-rank adapters for unseen English dialects, has also demonstrated the power of adaptive methods by addressing robustness across diverse linguistic scenarios.

Building on these advancements, recent work has sought to further enhance adaptability through more modular and context-sensitive designs. \cite{ostapenko2024modular} proposed a modular approach for LLMs by building and reusing a library of LoRAs. Unlike standard LoRA—which applies a single static low-rank update during inference—this modular method dynamically selects or combines specialized LoRA modules based on the task or context, thereby enabling a more flexible and responsive adaptation to diverse and evolving input distributions. In contrast, an alternative strategy employs pre-learned changeable masks \citep{sun2025text}. This approach involves training a set of masks—each tailored to specific tasks or data distributions—that are blended or selected during inference via a softmax-based selector informed by a task embedding or similar indicator. While these methods introduce dynamic adjustment, they come with some practical limitations, such as:
\begin{itemize}
    \item \textbf{Storage Overhead:} A large set of masks must be maintained, and these masks need to be loaded and blended during inference, increasing both memory and computational demands.
    \item \textbf{Adaptation Flexibility:} Even though blending introduces some variability, the masks are fine-tuned individually and not co-optimized for joint adaptation. This can lead to suboptimal performance when the masks are applied in a blended fashion.
    \item \textbf{Task Limitation:} Many of these approaches rely on task embeddings, which restrict adaptation to known task categories rather than capturing the broader context of the input batch.
\end{itemize}

Another line of research has focused on hypernetwork-based task conditioning and multi-task learning. As an early attempt, \citet{karimi2021parameter} proposed a shared hypernetwork framework for parameter‑efficient multi‑task fine‑tuning. Later, \citet{he2022hyperprompt} introduced Hyperprompt, a method that conditions transformers using hypernetwork‑generated prompts, while \citet{ivison2022hyperdecoders} presented hyperdecoders capable of producing instance‑specific decoders for multi‑task NLP. More recently, \citet{zheng2024dlo} proposed a strategy that dynamically adjusts model depth on a per‑sample basis by generating context‑dependent low‑rank updates via a hypernetwork, and \citet{schug2024attention} reinterpreted the attention mechanism as a hypernetwork, embedding dynamic parameter generation in the model’s core architecture. Lastly, \citet{ortiz2024hyperloader} integrated hypernetwork‑based LoRA and adapter layers into multi‑task transformers for sequence labeling, while \citet{lv2024hyperlora} leveraged constrained low‑rank adapter generation to achieve efficient cross‑task generalization.

Recent research has also explored using task description embeddings derived from powerful models, such as GPT-4, to generate low-rank updates on a per-sample basis \citep{charakorn2024instant}. This method generates the low-rank parameters for each input based on a task representation. While effective in certain scenarios, this per-sample adaptation may not fully exploit the shared context available when processing batches of similar inputs. ChameleonLLM differentiates itself by using batch-level context for adaptation. By clustering inputs and computing aggregated statistics, our method generates low-rank updates that reflect the collective characteristics of the batch. Aggregating across a batch helps mitigate the noise or outlier effects in single-sample adaptation. A uniform adaptation across a cluster of similar inputs can lead to more coherent and consistent outputs. By not being limited to task-specific embeddings, our method can adapt to a broader range of scenarios, including open-domain and instruction-based tasks. Methods that use task embeddings typically generate per-sample low-rank updates based on a fixed task identifier. In contrast:
\begin{itemize}
    \item \textbf{Batch vs. Sample-Level Adaptation:} Our method leverages batch-level statistics, leading to a more coherent adaptation that benefits from the collective context of multiple samples.
    \item \textbf{Flexibility in Unstructured Environments:} Task embeddings require clear task boundaries, whereas our context-based approach naturally adapts to real-world data.
    \item \textbf{Out-of-Sample Robustness:} Aggregating over a batch reduces the over-fitting sensitivity of training, an important advantage when dealing with heterogeneous or noisy datasets.
\end{itemize}

To sum up, in contrast to per-sample or depth-focused adaptation methods, ChameleonLLM exploits batch-level context to generate adaptive low-rank updates. Rather than adapting each sample independently, we cluster similar inputs and compute aggregated token embedding statistics across the batch; these statistics then drive our hyper-network to generate low-rank updates that reflect the collective context. ChameleonLLM achieves a coherent and context-aware adaptation that naturally overcomes the limitations of fixed, per-sample updates and the storage overhead associated with large sets of pre-learned masks, such as the impact of noisy or outlier samples, and the storage overhead associated with maintaining pre-learned masks. To conclude, while traditional methods like static LoRA and changeable masks offer efficient adaptation strategies, the recent surge in dynamic approaches demonstrates a growing consensus: incorporating context—whether at the sample or task level—can significantly enhance the adaptability and robustness of LLMs. Our work builds on previous research by uniquely leveraging batch-level clustering to generate context-aware low-rank updates, thereby improving the hypernetwork-based adaptation using aggregated batch statistics.

\section{Methodology}

In this work, we introduce \textbf{ChameleonLLM}, a dynamic adaptation framework leveraging batch-aware clustering and hyper-network–driven low-rank adaptations to efficiently tailor the model to the input context. ChameleonLLM can be built upon any pre-trained causal language model (e.g., GPT‑2) by augmenting the conventional transformer layers and language modeling head with low-rank adaptation modules. The innovation lies in the dynamic generation of low-rank parameters based on batch-aware statistics. This process involves two major components: (1) batch-aware clustering and (2) adaptive low-rank update generation. The central hypothesis is that inputs arriving in a batch often exhibit significant similarity---whether semantic, syntactic, or stylistic. By grouping similar inputs, we can extract shared context and exploit common features in the batch. This grouping not only aids in noise reduction by averaging but also ensures that the low-rank updates are tailored to the collective characteristics of the batch rather than isolated samples. Using the normalized token embeddings, we perform k‑means clustering. The number of clusters is chosen based on the batch size to ensure each cluster is large enough to provide robust statistics while capturing meaningful variations between groups. The clustering algorithm minimizes the Euclidean distance between points and cluster centroids, and through iterative refinement, it assigns each input to a cluster that best represents its features. Once the clusters are defined, the batch is reconstructed such that each mini-batch contains inputs from the same cluster. This reorganization is critical because it ensures that subsequent low-rank adaptations are computed on a homogenous set of inputs. 

The core idea of dynamic adaptation is to generate low-rank modifications specific to the current batch's context. This work proposes a more flexible alternative to LoRA that uses a hyper-network. Instead of applying a fixed set of low-rank updates (as in standard LoRA), ChameleonLLM employs a hyper-network to produce these parameters on-the-fly. The hyper-network takes as input the mean of the token embeddings from the entire batch, processes this aggregated statistic through several fully connected layers with non-linear activations, and outputs the parameters for the low-rank update. This on-the-fly generation allows the model to adjust to the specifics of the batch dynamically, capturing nuances that a static update might miss. The overall architecture maintains the integrity of the pre-trained model while introducing minimal additional parameters. The transformer layers are wrapped with LoRA modules that operate similarly to conventional implementations, ensuring that most of the model’s capacity remains unchanged. For the LM head, the hyper-network variant ensures that adaptation is context-dependent. During inference, the model first processes the batch to compute the necessary embeddings to cluster the inputs and then generates a custom low-rank update for the LM head based on the cluster's aggregated statistics. The final output is produced by applying this adapted LM head to the last hidden states of the transformer.

\subsection{Implementation}

To leverage the inherent similarity between samples, we first cluster the normalized token embeddings using the k-means algorithm.  This ensures that mini-batches are contextually coherent and provide reliable aggregate statistics for conditioning the hyper-network. The number of clusters is determined dynamically (typically by dividing the dataset size by the batch size) to ensure each cluster is sufficiently large to provide reliable aggregate statistics while preserving the similarity among examples. Once clustering is complete, we construct the training and validation data loaders that generate mini-batches composed of homogeneous examples from the same cluster. This \emph{clustered batching} strategy is critical because it ensures that the subsequent dynamic adaptation via the hyper-network is conditioned on inputs with similar features, enhancing the adaptability and robustness of the generated low-rank updates. By integrating detailed data preprocessing, batch-aware clustering, and dynamic low-rank adaptation, our methodology efficiently leverages the strengths of a pre-trained transformer backbone while adapting to the specific context of each batch.

In our architecture, the transformer layers are augmented with the standard LoRA modules, while the LM head is replaced with a hyper-network LoRA module. Designed specifically for the language modeling (LM) head, this module dynamically predicts update matrices based on the aggregated token embeddings from the current batch. The hyper-network processes the mean token embedding through several fully connected layers with non-linear activations (e.g., ReLU) and dropout, ultimately outputting the low-rank matrices (denoted as A and/or B). Initialization is performed using Xavier and small-scale normal initialization. A dedicated procedure handles wrapping by freezing the original model parameters and applying the appropriate LoRA modifications. This ensures that only the LoRA and hyper-network components are trainable, thereby preserving the representational capacity of the pre-trained model. The model is trained using a standard next-token prediction objective with cross-entropy loss. During each training iteration, the following steps are executed:
\begin{enumerate}
    \item \textbf{Transformer Processing}: The input batch of clustered examples, is passed through the frozen transformer layers augmented with LoRA, yielding hidden states for each token.
    
    \item \textbf{Dynamic LM Head Adaptation}: The last hidden state is passed to the LM head. Simultaneously, the LM head receives the aggregated token embeddings (computed as the mean over the batch) as input, allowing the hyper-network to generate dynamic low-rank updates.
    
    \item \textbf{Prediction and Loss Computation}: The adapted LM head produces logits for next-token prediction, which are compared against the target tokens to compute the cross-entropy loss.
    
    \item \textbf{Optimization}: Gradients are computed using the AdamW optimizer, and only the LoRA and hyper-network parameters are updated.
\end{enumerate}

\section{Experiments}

For our experiments, we have used WikiText-2 \citep{merity2016pointer} and Alpaca \citep{alpaca} datasets, providing meaningful benchmarks due to their diverse and natural language texts. The datasets are split into training and validation sets. Each text sample is tokenized using a pre-trained GPT-2 tokenizer by truncating to a maximum length and padding to ensure uniform sequence lengths. LM input token embeddings are computed for every example, and normalized to facilitate robust clustering. For the Alpaca dataset, we calculate the token embeddings from the instruction part of the input prompts. A crucial part of ChameleonLLM is creating data loaders that reflect the clustered nature of the inputs. Using the precomputed token embeddings, we apply k‑means clustering with the number of clusters based on the desired batch size and overall dataset size. After clustering, indices are grouped so each mini-batch contains examples from a single cluster. This ensures that each batch is contextually coherent. We implement a custom data loader that leverages these clusters as batch samplers. The data loader returns batches fed into the model during training and evaluation. The evaluation is performed at the end of each epoch on the validation set, where batches are processed through the clustering pipeline, and the average loss is computed. This systematic evaluation ensures that our comparisons are fair and reflect real-world performance. All codes for ChameleonLLM are \href{https://anonymous.4open.science/r/ChamaleonLLM/}{open-sourced} to ensure the reproducibility of experimental results.

\begin{table}[htbp]
  \centering
  \caption{Comparison of Low-Rank Adaptation Regimes on WikiText-2 Dataset}
  \label{tab:gpt2_comparison_wikitext}
  \begin{tabular}{lccccc}
    \toprule
    Adaptation Regime & Parameters & Training Loss & Validation Loss & Val. Perplexity \\
    \midrule
    Unadapted GPT-2        & 124,439,808       & 13.8144 & 13.7876   & 972,500 \\
    Traditional LoRA        & 204,100  & 0.5504  & 0.5023   & 1.6525 \\
    \textbf{ChameleonLLM}       & 6,786,596 & \textbf{0.2359}  & \textbf{0.3753}    & \textbf{1.4554} \\
    \bottomrule
  \end{tabular}
\end{table}

\begin{table}[htbp]
  \centering
  \begin{threeparttable}
    \caption{Comparison of Low-Rank Adaptation Regimes on Alpaca Dataset}
    \label{tab:gpt2_comparison_alpaca}
    \begin{tabular}{lccccc}
      \toprule
      Adaptation Regime & Parameters & Training Loss & Validation Loss & Val. Perplexity \\
      \midrule
      Unadapted GPT-2         & 124,439,808 & 13.3881 & 13.3881 & 652,166 \\
      Traditional LoRA        & 204,100     & 0.1254  & 0.1527  & 1.1650 \\
      \textbf{ChameleonLLM\tnote{*}} & 6,786,596 & \textbf{0.0810} & \textbf{0.1082} & \textbf{1.1143} \\
      \bottomrule
    \end{tabular}
    \begin{tablenotes}
      \footnotesize
      \item[*] Token embeddings for adaptation are calculated from the instruction part of the prompt only.
    \end{tablenotes}
  \end{threeparttable}
\end{table}

\begin{itemize}
    \item \textbf{Traditional LoRA Fine-Tuning:} Here, all transformer layers and the LM head use the static LoRA adaptation. Standard cross-entropy loss (with appropriate masking for padding tokens) is computed, and the low-rank parameters are optimized using AdamW optimizer.
    \item \textbf{ChameleonLLM Fine-Tuning:} In this regime, while the transformer layers continue to use static LoRA modules, the LM head is adapted by a hyper-network LoRA module that generates low-rank update parameters based on the mean token embeddings of the batch. The loss is computed on the LM head outputs, and optimized similarly.
\end{itemize}

Experimental results demonstrate that ChameleonLLM significantly improves over the traditional LoRA approach, consistently achieving a lower average validation loss on both datasets, and suggest that dynamic adaptation based on batch statistics leads to better convergence and generalization. As shown in Table~\ref{tab:gpt2_comparison_wikitext}, ChameleonLLM achieves a validation loss reduction of approximately \textbf{25\%} compared to traditional LoRA, along with a perplexity improvement of roughly 12\%. Table~\ref{tab:gpt2_comparison_alpaca} further indicates that it reduces the validation loss by almost \textbf{30\%} on instruction fine-tuning. These improvements underscore the effectiveness of the batch-level context adaptation strategy and can largely be attributed to the exploitation of batch-level context. By averaging token embeddings over each batch, the hyper-network input becomes less sensitive to individual outliers, leading to robust low-rank parameter generation. Clustering groups together semantically or stylistically similar inputs allows the hyper-network to extract stronger, more relevant signals for adaptation. 

Although ChameleonLLM has more trainable parameters than LoRA, this increase is justified by many practical benefits, as it dynamically generates low-rank updates on-the-fly rather than training and storing separate adaptation matrices for each sample or task. Avoiding the storage of numerous pre-learned matrices significantly reduces memory and processing overheads during inference. The additional parameters enable context-aware, batch-level adaptation, leading to better performance. Despite it also introduces some extra computational steps during inference, the overhead is modest compared to the overall inference time. The trade-off between increased computation and improved performance is highly favorable. Our approach is not tied to pre-defined task categories and adapts to the nuances in the data at inference time, offering a versatile solution for open-domain tasks.
    
\section{Conclusion}

In this paper, we presented ChameleonLLM---a novel framework that enables inference-time adaptation of large language models through batch-aware clustering and dynamic low-rank parameter generation. By clustering similar inputs and using a hyper-network to generate low-rank updates based on the aggregated batch statistics, our method provides a flexible and efficient alternative to static fine-tuning approaches such as traditional LoRA or uniform mask blending. Our extensive experiments on WikiText-2 and Alpaca demonstrate that ChameleonLLM achieves lower validation loss and perplexity than baseline methods with dynamic batch-aware contextual adaptation. The proposed framework reduces the need for storing multiple expert masks and adapts to diverse input distributions in real-time, making it highly suitable for modern, dynamic inference environments.

\bibliography{iclr2025_conference}

\begin{thebibliography}{17}
\providecommand{\natexlab}[1]{#1}
\providecommand{\url}[1]{\texttt{#1}}
\expandafter\ifx\csname urlstyle\endcsname\relax
  \providecommand{\doi}[1]{doi: #1}\else
  \providecommand{\doi}{doi: \begingroup \urlstyle{rm}\Url}\fi

\bibitem[Charakorn et~al.(2024)Charakorn, Cetin, Tang, and Lange]{charakorn2024instant}
Rujikorn Charakorn, Edoardo Cetin, Yujin Tang, and Robert~Tjarko Lange.
\newblock Instant transformer adaption via hyperlora.
\newblock In \emph{Adaptive Foundation Models: Evolving AI for Personalized and Efficient Learning}, 2024.

\bibitem[Chen et~al.(2024)Chen, Chen, Cheng, and Chen]{chen2024multiplicative}
Xuxi Chen, Tianlong Chen, Yu~Cheng, and Weizhu Chen.
\newblock One is not enough: Parameter-efficient fine-tuning with multiplicative sparse factorization.
\newblock \emph{IEEE Journal of Selected Topics in Signal Processing}, Sep 2024.
\newblock To appear.

\bibitem[Ha et~al.(2016)Ha, Dai, and Le]{ha2016hypernetworks}
David Ha, Andrew Dai, and Quoc~V Le.
\newblock Hypernetworks.
\newblock \emph{arXiv preprint arXiv:1609.09106}, 2016.

\bibitem[He et~al.(2022)He, Zheng, Tay, Gupta, Du, Aribandi, Zhao, Li, Chen, Metzler, et~al.]{he2022hyperprompt}
Yun He, Steven Zheng, Yi~Tay, Jai Gupta, Yu~Du, Vamsi Aribandi, Zhe Zhao, YaGuang Li, Zhao Chen, Donald Metzler, et~al.
\newblock Hyperprompt: Prompt-based task-conditioning of transformers.
\newblock In \emph{Proceedings of the International Conference on Machine Learning}, pp.\  8678--8690. PMLR, 2022.

\bibitem[Hu et~al.(2021)Hu, Shen, Wallis, Allen-Zhu, Li, Wang, Wang, and Chen]{hu2021lora}
Edward~J Hu, Yelong Shen, Phillip Wallis, Zeyuan Allen-Zhu, Yuanzhi Li, Shean Wang, Lu~Wang, and Weizhu Chen.
\newblock Lora: Low-rank adaptation of large language models.
\newblock \emph{arXiv preprint arXiv:2106.09685}, 2021.

\bibitem[Ivison \& Peters(2022)Ivison and Peters]{ivison2022hyperdecoders}
Hamish Ivison and Matthew~E Peters.
\newblock Hyperdecoders: Instance-specific decoders for multi-task nlp.
\newblock \emph{arXiv preprint arXiv:2203.08304}, 2022.

\bibitem[Kim et~al.(2024)Kim, Sasaki, Hoshino, and Honda]{kim2024single}
Hwichan Kim, Shota Sasaki, Sho Hoshino, and Ukyo Honda.
\newblock A single linear layer yields task-adapted low-rank matrices.
\newblock \emph{arXiv preprint arXiv:2403.14946}, 2024.

\bibitem[Lv et~al.(2024)Lv, Li, Zhang, Chen, Qi, Zhang, and Zheng]{lv2024hyperlora}
Chuancheng Lv, Lei Li, Shitou Zhang, Gang Chen, Fanchao Qi, Ningyu Zhang, and Hai-Tao Zheng.
\newblock Hyperlora: Efficient cross-task generalization via constrained low-rank adapters generation.
\newblock In \emph{ACL Findings}, 2024.
\newblock URL \url{https://openreview.net/forum?id=xa4GYUSvhW}.

\bibitem[Mahabadi et~al.(2021)Mahabadi, Ruder, Dehghani, and Henderson]{karimi2021parameter}
Rabeeh~Karimi Mahabadi, Sebastian Ruder, Mostafa Dehghani, and James Henderson.
\newblock Parameter-efficient multi-task fine-tuning for transformers via shared hypernetworks.
\newblock \emph{arXiv preprint arXiv:2106.04489}, 2021.

\bibitem[Merity et~al.(2016)Merity, Xiong, Bradbury, and Socher]{merity2016pointer}
Stephen Merity, Caiming Xiong, James Bradbury, and Richard Socher.
\newblock Pointer sentinel mixture models, 2016.

\bibitem[Ortiz-Barajas et~al.(2024)Ortiz-Barajas, Gomez-Adorno, and Solorio]{ortiz2024hyperloader}
Jesus-German Ortiz-Barajas, Helena Gomez-Adorno, and Thamar Solorio.
\newblock Hyperloader: Integrating hypernetwork-based lora and adapter layers into multi-task transformers for sequence labelling.
\newblock \emph{arXiv preprint arXiv:2407.01411}, 2024.

\bibitem[Ostapenko et~al.(2024)Ostapenko, Su, Ponti, Charlin, Le~Roux, Pereira, Caccia, and Sordoni]{ostapenko2024modular}
Oleksiy Ostapenko, Zhan Su, Edoardo~Maria Ponti, Laurent Charlin, Nicolas Le~Roux, Matheus Pereira, Lucas Caccia, and Alessandro Sordoni.
\newblock Towards modular llms by building and reusing a library of loras.
\newblock \emph{arXiv preprint arXiv:2405.11157}, 2024.

\bibitem[Schug et~al.(2024)Schug, Kobayashi, Akram, Sacramento, and Pascanu]{schug2024attention}
Simon Schug, Seijin Kobayashi, Yassir Akram, Jo{\~a}o Sacramento, and Razvan Pascanu.
\newblock Attention as a hypernetwork.
\newblock \emph{arXiv preprint arXiv:2406.05816}, 2024.

\bibitem[Sun et~al.(2025)Sun, Cetin, and Tang]{sun2025text}
Qi~Sun, Edoardo Cetin, and Yujin Tang.
\newblock $\text {Transformer}^ 2$: Self-adaptive llms.
\newblock \emph{arXiv preprint arXiv:2501.06252}, 2025.

\bibitem[Tan et~al.(2024)Tan, Dong, Zhao, Peng, Cheng, and Chen]{zheng2024dlo}
Zhen Tan, Daize Dong, Xinyu Zhao, Jie Peng, Yu~Cheng, and Tianlong Chen.
\newblock Dlo: Dynamic layer operation for efficient vertical scaling of llms, 2024.
\newblock ArXiv preprint arXiv:2407.11030.

\bibitem[Taori et~al.(2023)Taori, Gulrajani, Zhang, Dubois, Li, Guestrin, Liang, and Hashimoto]{alpaca}
Rohan Taori, Ishaan Gulrajani, Tianyi Zhang, Yann Dubois, Xuechen Li, Carlos Guestrin, Percy Liang, and Tatsunori~B. Hashimoto.
\newblock Stanford alpaca: An instruction-following llama model.
\newblock \url{https://github.com/tatsu-lab/stanford_alpaca}, 2023.

\bibitem[Xiao et~al.(2023)Xiao, Held, Liu, and Yang]{xiao2023task}
Zedian Xiao, William Held, Yanchen Liu, and Diyi Yang.
\newblock Task-agnostic low-rank adapters for unseen english dialects.
\newblock In \emph{Proceedings of the 2023 Conference on Empirical Methods in Natural Language Processing}, pp.\  7857--7870. Association for Computational Linguistics, 2023.
\newblock \doi{10.18653/v1/2023.emnlp-main.487}.
\newblock URL \url{https://aclanthology.org/2023.emnlp-main.487}.

\end{thebibliography}
\bibliographystyle{iclr2025_conference}

\end{document}